# Algorithm for Missing Values Imputation in Categorical Data with Use of Association Rules

Jiří Kaiser[1]

[1] Czech Technical University in Prague – Faculty of Civil Engineering / Department of Applied Informatics, Prague, Czech Republic
Email: jiri.kaiser@fsv.cvut.cz

*Abstract*—**This paper presents new algorithm for missing values imputation in categorical data. The algorithm is based on using association rules and is presented in three variants. Experimental shows better accuracy of missing values imputation using new algorithm then using most common attribute value.**

## I. INTRODUCTION

The need to address the problems with missing values in the data occurs in the preparation of data for analyses. The first possible solution of this problem is reducing the data set. This way of handling missing values is still commonly used in practice [1] but may cause significant loss of usable data. The other possible solution is missing values imputation. Selection of missing values imputation method must be done with regard to the structure of the data set. Many described methods can be used for missing values imputation in numerical data, e.g.: mean substitution [2], linear regression [2], neural networks [3] and nearest neighbor approach [4]. The commonly used method for missing values imputation in categorical data is to substitute missing values of each attribute by the most common value of the attribute [5]. The disadvantage of this method is that it does not consider dependencies among attributes values. This paper describes three variants of new algorithm for missing values imputation with use of association rules and presents results of tests.

## II. GENERATING ASSOCIATION RULES

An association rule takes the following form [6]:

$$IF\ A = 1\ AND\ B = 1\ THEN\ C = 1, \quad (1)$$

where A, B, and C are variables and

$$p = p\ (\ C = 1\ |\ A = 1,\ B = 1\ ). \quad (2)$$

i.e., the conditional probability that C = 1 given that A = 1 and B = 1. The conditional probability p is referred to as the "confidence" of the rule, and p (A = 1, B = 1, C = 1) is referred to as the "support". The support can be used as a constraint of minimum count of cases supporting association rule. The "If" part of the rule is often called antecedent and the "then" part is often called consequent [7].

There are several ways of generating association rules, e.g.: apriori algorithm [8], [6], [9], GUHA method [8] and other algorithms derived from apriori algorithm [10].

## III. OBTAINING COMPLETE DATA SET FOR GENERATING ASSOCIATION RULES

Algorithms for association rules generation are usually unable to handle missing values. There were two possibilities of getting data set for association rules generation. The first way to get complete data set for association rules generation was reducing the data set. The second way was to handle missing values as special values [5]. This was practically done by replacement of all missing values by value "MISSING". The second way was chosen for the algorithm because of advantage in possibility of handling data sets with high amount of missing values.

## IV. VARIANTS OF THE ALGORITHM

The designed algorithm uses two data sets. The first data set was called "training data set" and contains data for generation of association rules. The second data set was called "data set for missing values imputation" and contained data with imputed values. All variants of the algorithm assume that at the start of the algorithm both data sets have the same content as the data set with missing values.

### A. The First Variant

One way of using association rules for missing values imputation is following:
1. In the training data set fill all missing values by the special value. In further steps are these values called "MISSING".
2. Generate association rules from the training data set.
3. From the list of association rules remove rules with support lower than required.
4. From the list of association rules remove rules with consequent that is combination longer than 1.
5. From the list of association rules remove rules with consequent containing value "MISSING".
6. Sort association rules by confidence in descending order
7. For each value "MISSING" in the training data set pass through the list of association rules until the suitable rule is found or until the end of the association rules list is reached. Suitable rule is the one meeting following requirements:

- consequent contains the value of attribute whose value is being searched,
- antecedent corresponds to values of other given case attributes.
8. If suitable association rule was found, fill the missing value in the data set for missing values imputation by value in the consequent of the association rule.

*B. The Second Variant*

The first variant of the algorithm was unable to impute missing values if there were no suitable association rule. This problem was handled by combining association rules and most common attribute value. First was used association rules method and if there was no suitable association rule then the most common attribute value was used.

The second variant of designed algorithm:
1. In the training data set fill all missing values by the special value. In further steps are these values called "MISSING".
2. Generate association rules from the training data set.
3. From the list of association rules remove rules with support lower than required.
4. From the list of association rules remove rules with consequent that is combination longer than 1.
5. From the list of association rules remove rules with consequent containing value "MISSING".
6. For each attribute find the most common value (except the value "MISSING").
7. Sort association rules by confidence in descending order.
8. For each value "MISSING" in the training data set pass through the list of association rules until the suitable rule is found or until the end of the association rules list is reached. Suitable rule is the one meeting following requirements:
   - consequent contains the value of attribute whose value is being searched,
   - antecedent corresponds to values of other given case attributes.
9. If suitable association rule was found, fill missing value in the data set for missing values imputation by value in consequent of the association rule. Else fill missing value in the data set for missing values imputation by the most common attribute value (except the value "MISSING").

*C. The Third Variant*

The third variant of the algorithm was designed to improve missing values imputation accuracy using better combination of association rules and the most common attribute value method. Only association rules with confidence not lower than the relative frequency of occurrence of the most common value of the attribute were used. For the imputation of the rest of missing values was used the most common attribute value method. This approach was based on fact that the most common attribute value can be handled as a special zero attribute rule [11], [12]. The antecedent of this rule contains no attributes and consequent contains one attribute. Support and confidence of this rule is equal to the relative frequency of the most common value of the attribute.

The third variant of designed algorithm:
1. In the training data set fill all missing values by the special value. In further steps are these values called "MISSING".
2. For each attribute find the most common value (except the value "MISSING") and compute its relative frequency of occurrence.
3. Generate association rules from the training data set.
4. From the list of association rules remove rules with support lower than required.
5. From the list of association rules remove rules with consequent that is combination longer than 1.
6. From the list of association rules remove rules with consequent containing value "MISSING".
7. Sort association rules by confidence in descending order.
8. For each value "MISSING" in the training data set pass through the list of association rules until the suitable rule is found or until the end of the association rules list is reached. Suitable rule is the one meeting following requirements:
   - consequent contains the value of attribute whose value is being searched,
   - antecedent corresponds to values of other given case attributes,
   - confidence of the association rule is not lower than the relative frequency of the most common value (except the value "MISSING") whose value is being searched.
9. If suitable association rule was found, fill missing value in the data set for missing values imputation by value in consequent of the association rule. Else fill missing value in the data set for missing values imputation by the most common attribute value (except the value "MISSING").

## V. ALGORITHM TESTS

*A. Testing Procedure*

In real situation missing values are unknown and it is not possible to verify if missing value was estimated correctly. For testing purposes were missing values generated randomly in complete data sets.

Following procedure for algorithm testing was used:
1. The incomplete data set was created by generation of required amount of missing values in complete data set.
2. The missing value imputation method was applied.
3. Obtained data set was compared with original data set and amount of missing values estimated incorrectly was counted.

Steps 2 and 3 were done for the third variant of designed algorithm and for the most common attribute value method and gained results were compared. Tests were done for 1, 2, 5, 10, 20, 40 and 70% of missing values.

*B. Used Data Sets*

The algorithm was tested on three data sets. The first one contained categorical attributes from data set

"Physical examination" from Stulong study[1]. Used data set contained 5 attributes, 10556 cases and 52780 values. The second and the third data set contained 3 attributes, 2000 cases and 6000 values. The second data set had first attribute values generated using pseudo-random numbers generator and values of other attributes were directly derived from the values of the first attribute. All attributes values of the third data set were generated randomly. The generated values in the second and the third data set had discrete uniform distribution.

Association rules were generated with use of Weka software [13].

*C. Tests Results*

The third variant of the algorithm was tested. For each required amount of missing values were five or more tests done. Following three tables show average % of incorrectly estimated missing values. Results are shown for the third variant of the algorithm and for the most common attribute value method.

Tables show that the best results were reached on the second data set and the worst on the third data set. The new algorithm got better results than the most common attribute value method in all tests on the first and the second data sets. Tests on the third data set show that missing values imputation accuracy was similar to accuracy of the most common attribute value method.

## VI. DISCUSSION

The problem of testing missing values imputation methods is in generation of missing values for testing procedure. Generated missing values for testing purposes had a discrete uniform distribution. In real situation missing values may have many reasons and may not have discrete uniform distribution.

Another question is the setting of minimum required support of association rules. Adjusting support may cause getting different results. For testing purposes was condition of minimum required support set to 1 case. Setting requirement of association rules support higher may cause getting higher missing imputation accuracy but using too high requirement of association rules may cause small amount of usable association rules and decrease missing values imputation accuracy significantly.

The results of missing values imputation depend on association rules order in the list of association rules. If there are more association rules with the same confidence, the imputed value may depend on the order of these association rules in the list. Order of association rules with the same confidence is dependent on used algorithm for generation of association rules.

[1] The study (STULONG) was realized at the 2nd Department of Medicine, 1st Faculty of Medicine of Charles University and Charles University Hospital, U nemocnice 2, Prague 2 (head. Prof. M. Aschermann, MD, SDr, FESC), under the supervision of Prof. F. Boudík, MD, ScD, with collaboration of M. Tomečková, MD, PhD and Ass. Prof. J. Bultas, MD, PhD. The data were transferred to the electronic form by the European Centre of Medical Informatics, Statistics and Epidemiology of Charles University and Academy of Sciences (head. Prof. RNDr. J. Zvárová, DrSc). The data resource is on the web pages http://euromise.vse.cz/challenge2003. At present time the data analysis is supported by the grant of the Ministry of Education CR Nr LN 00B 107.

TABLE I.
RESULTS OF TESTS ON THE FIRST DATA SET

| Average % of incorrectly estimated missing values using | % of Missing Values | | | | | | |
|---|---|---|---|---|---|---|---|
| | 1 | 2 | 5 | 10 | 20 | 40 | 70 |
| assoc. rules algorithm | 19,8 | 20,9 | 21,2 | 21,6 | 21,7 | 22,3 | 22,7 |
| most common attrib. value | 22,1 | 23,5 | 23,7 | 23,7 | 23,7 | 23,7 | 23,7 |

TABLE II.
RESULTS OF TESTS ON THE SECOND DATA SET

| Average % of incorrectly estimated missing values using | % of Missing Values | | | | | | |
|---|---|---|---|---|---|---|---|
| | 1 | 2 | 5 | 10 | 20 | 40 | 70 |
| assoc. rules algorithm | 0 | 0 | 0,4 | 0,8 | 3,3 | 10,8 | 32,6 |
| most common attrib. value | 70 | 67,2 | 64,9 | 67,2 | 67,2 | 66,7 | 66,6 |

TABLE III.
RESULTS OF TESTS ON THE THIRD DATA SET

| Average % of incorrectly estimated missing values using | % of Missing Values | | | | | | |
|---|---|---|---|---|---|---|---|
| | 1 | 2 | 5 | 10 | 20 | 40 | 70 |
| assoc. rules algorithm | 65,1 | 66,0 | 65,2 | 65,5 | 66,5 | 66,0 | 66,4 |
| most common attrib. value | 67,0 | 66,8 | 67,4 | 66,8 | 67,7 | 66,6 | 66,7 |

In the third variant of the algorithm may be generated only association rules with confidence not lower than the relative frequency of occurrence of the most common value (except the value "MISSING"). The most common value and its relative frequency of occurrence is different for each attribute but algorithms for generating association rules usually allows to set condition of minimum confidence of association rules only in general. In this case the most common value (except the value "MISSING") with lowest relative frequency of occurrence must be used.

## CONCLUSIONS

The tests results confirmed expectation that the new missing values imputation algorithm using association rules has increasing missing values imputation accuracy with increasing dependence among attribute values in the data set. In the data set with independent attributes is missing values imputation accuracy similar to accuracy of the most common attribute value method. As a future

ACEEE

work, it is required to search for the way of setting minimum required support of association rules to get the best accuracy. Another goal should be an optimization of association rules generation for missing values imputation purpose.

\